\documentclass[11pt]{article}

\usepackage[T1]{fontenc}
\usepackage[utf8]{inputenc}
\usepackage{lmodern}
\usepackage[margin=1in]{geometry}
\usepackage{microtype}
\usepackage{amsmath}
\usepackage{graphicx}
\usepackage{booktabs}
\usepackage{tabularx}
\usepackage{array}
\usepackage{float}
\usepackage[round,authoryear]{natbib}
\setcitestyle{aysep={}}
\usepackage[font=small,labelfont=bf,labelsep=period]{caption}
\usepackage[hidelinks]{hyperref}
\usepackage{xurl}

\newif\ifdraft
\draftfalse

\newcolumntype{L}{>{\raggedright\arraybackslash}X}

\title{\textbf{Validity Without Ground Truth: What Stated-Preference Economics Offers the Evaluation of Language Models}}

\author{%
  Daniel Robert Kling Alexander\thanks{Daniel Robert Kling Alexander is a 2026 M.S.\ graduate from the University of Michigan School of Information. \url{https://github.com/drkalexander1}}
  \and
  Catherine Louise Kling\thanks{Catherine Louise Kling is a Tisch University Professor, SC Johnson College of Business, Cornell University. \url{https://business.cornell.edu/profiles/clk228/}}%
}

\newcommand{\availability}{%
  \thanks{\textbf{Data and code availability.} Code, prompts, frozen predictions, and analysis scripts are at \url{https://github.com/drkalexander1/llm-valuation-evals}. AI coding assistants (Cursor and Claude) were used to write code, check results, and draft text; the repository README describes their roles. The authors designed the study and take full responsibility for its content.}}

\ifdraft
  \date{\textit{Draft, 6 October 2026. Do Not Cite Without Author Permission}\availability}
\else
  \date{October 2026\availability}
\fi

\begin{document}

\maketitle

\begin{abstract}
\noindent Many of the questions now put to large language models have no correct answer to score against: what a policy is worth, which option a user should choose, how to weigh competing values. Stated-preference economics has faced this problem for decades. It judges survey responses without knowing the true value, through a framework of validity and related concepts: content, construct, and criterion validity, reliability, incentive compatibility, and consequentiality. We argue that this framework is a general method for evaluating language models, and we set out what each concept means for LLM evaluation. We demonstrate the approach using a published water-quality stated preference economic valuation survey \citep{vossler2023} administered to six models. In this economic application, the validity tests take the form of predictions from economic theory: demand should slope down, and willingness to pay should respond to the scope of the good and to income. The tests separate the models sharply. Two older models fail the most basic test at a household income level of \$75,000, and the two newest pass every test of theoretical validity we can score, but diverge on convergent validity. Passing validity tests shows that a model's answers are coherent, not that they are correct.
\end{abstract}

\section{Introduction}

Evaluations of large language models (LLMs) mostly score answers against a key. A growing share of what people ask these systems has no key: whether a purchase or a ballot measure is worth its cost, which of several options suits a user, how to trade off competing values. Researchers and firms also increasingly use LLMs in place of survey respondents. For such questions, accuracy is undefined, and matching human answers is a weak substitute, because human answers are themselves uncertain and a model could reason well without reproducing them. How, then, can we tell whether a model's answers are any good?

Economists who value goods that are not sold in markets have faced exactly this problem. Stated-preference (SP) surveys ask people how they would vote or choose when offered a hypothetical policy at a cost, and no market price exists to check the answers against. The field's response was to develop a framework of validity concepts for judging whether responses are credible without knowing the true value \citep{mitchell1989,kling2012,bishop2019}. \emph{Content validity} asks whether the instrument asks the right question clearly. \emph{Criterion validity} compares responses with a measure closer to the truth. \emph{Construct validity} asks whether responses agree with other measures of the same thing (\emph{convergent validity}) and move as theory predicts (\emph{theoretical validity}). Later work added \emph{reliability}, whether the answer survives repetition; \emph{incentive compatibility}, whether the question format rewards truthful answers \citep{carson2007}; and \emph{consequentiality}, whether respondents believe the survey matters \citep{carson2014,vossler2012}.

Our contribution is to recognize that this framework applies to LLM evaluation in general, not only to economic questions. These are general principles for judging answers that cannot be checked against the truth. In a valuation study, they happen to be expressed through consistency with economic theory: demand should slope down, and value should respond to the scope of the good and to income. In other domains, they would be expressed through that domain's theory, accepted practice, or common sense. Table~\ref{tab:framework} sets out what each concept means for LLM evaluation. Two of the mappings touch problems that AI evaluation already worries about. Consequentiality corresponds to evaluation awareness, whether a model behaves differently when it believes the stakes are real \citep{needham2025}. Incentive compatibility corresponds to sycophancy, whether a setup rewards telling the user what they want to hear \citep{sharma2024}.

The idea that AI evaluation is a measurement problem is not new. Work on measurement and fairness \citep{jacobs2021}, on benchmark validity \citep{raji2021}, and on construct validity in LLM benchmarks \citep{wallach2025,bean2025} draws on psychometrics to ask whether benchmarks measure what they claim. SP economics adds three things. It was built for questions with no right answer, while the benchmarks that work examines mostly score against one. It comes with concrete behavioral tests, such as scope tests, that can be run on a model directly. And its incentive concepts map onto live problems in LLM evaluation. Closer to our application, recent papers elicit economic preferences from LLMs, testing rationality under revealed preference \citep{chen2023} or comparing LLM-derived willingness to pay (WTP) with human benchmarks \citep{brand2026,reusens2026}. These studies concern private goods and score models on rationality axioms or closeness to human numbers. None uses the validity framework as a whole.

We demonstrate the framework with a published and well-vetted instrument, the water-quality referendum of \citet[\emph{PNAS}]{vossler2023}, given to six models from two developers. The application is an example, chosen because the instrument is well documented and its validity tests are well understood. The same approach applies to any setting where a model's answers can be tested against what should and should not change them.

\section{The validity framework for language models}

\begin{table}[ht]
\centering
\small
\renewcommand{\arraystretch}{1.25}
\begin{tabularx}{\textwidth}{@{}>{\raggedright\arraybackslash}p{2.6cm}LLL@{}}
\toprule
\textbf{Concept} & \textbf{Question in stated preference} & \textbf{Question for LLM evaluation} & \textbf{In our demonstration} \\
\midrule
Content validity & Does the instrument describe the good, payment, and decision clearly and credibly? & Does the task pose the question it claims to, and could the model be recalling the benchmark rather than answering it? & Published survey text, verbatim; the scale's published name withheld \\
Construct: theoretical & Do answers move as economic theory predicts? & Do answers respond to what should matter, in the right direction, and ignore what should not? & Demand slopes down; scope; income \\
Construct: convergent & Do different measures of the same value agree? & Do different prompts, formats, and scoring methods reach the same conclusion? & Two WTP estimators; referendum vs.\ open-ended; human estimates \\
Criterion validity & Do stated values match real payments? & Does what the model says match what it does when it acts? & Not assessed: the model never pays \\
Reliability & Does the answer survive repetition? & Are results stable across samples, reruns, and model versions? & Ten independent draws per question \\
Incentive compatibility & Does the format reward truthful answers? & Does the setup reward telling the user what they want to hear? & Referendum format; not tested \\
Consequentiality & Does the respondent believe the survey matters? & Does the model behave differently when it believes the stakes are real? & Not tested \\
\bottomrule
\end{tabularx}
\caption{Validity concepts from stated-preference economics \citep{mitchell1989,kling2012,bishop2019,carson2007,carson2014} and what each asks of an LLM evaluation.}
\label{tab:framework}
\end{table}

Three features of the mapping matter. First, theoretical validity transfers unchanged and does most of the work. It needs no true answer, only predictions about what should move a response. Scope tests are the SP literature's best-known example \citep{arrow1993,whitehead2016}: a larger good should be worth at least as much as a smaller one it contains. Second, reliability is cheap for a model, which can be sampled repeatedly, but it acquires a new dimension, stability across model versions. Third, the concepts that depend on a respondent with something at stake, criterion validity, incentive compatibility, and consequentiality, have no direct analogue. For LLMs they become open empirical questions rather than design properties.

One design choice sharpens the tests. We give the survey to the model as an adviser to a household, not as a respondent. A person whose values violate a scope test may simply have cognitive limits. An adviser who recommends paying less for a policy that delivers strictly more has made an error. Under an advisory framing, then, theoretical validity becomes a normative standard, and a failure is a mistake rather than a preference. The same holds for any evaluation of a model acting on a user's behalf.

\section{A demonstration: valuing water quality}

\textbf{Instrument.} \citet*{vossler2023} surveyed households in the Upper Mississippi, Ohio, and Tennessee River Basins to elicit economic values of potential improvements in ecological conditions related to water quality, measured on a six-level scale from Level 1 (natural state) to Level 6 (extreme degradation). Respondents voted in advisory referendums on policies that varied the type of improvement, the region, and a household tax paid annually for five years. We reproduce the survey's text: the level labels, the policy summary table, and the referendum question. We withhold the scale's published name, which respondents did not see either, to reduce the chance that a model recalls the study instead of answering it.

\textbf{Design.} Nine scenarios cross three improvements with three regions. A one-level improvement raises every area by one level; minimum Level 2 and minimum Level 3 raise every area below that level up to it. Minimum Level 2 therefore delivers everything minimum Level 3 does and strictly more. The regions are the household's local watershed, a non-local watershed of the same size that does not contain the home, and the full study region, which is roughly 25 times larger and contains the home. The baseline is held fixed across regions. The model is told it is advising a four-person household with a stated income, may reason freely, and must end with a Yes or No vote. Each scenario is posed at ten annual tax levels from \$20 to \$3,000, at household incomes of \$35,000, \$75,000 (about the average in the human sample), and \$200,000, with ten independent draws each. The six models are Claude Haiku 4.5, Claude Sonnet 4.5, and Claude Sonnet 5 from Anthropic, and GPT-4o mini, GPT-4o, and GPT-5.6 Terra from OpenAI, spanning size and generation. A first, exploratory round (19--20 September 2026) ran the watershed scenarios at \$75,000. Before a second round (27--28 September) ran the remaining cells, all nine scenarios at \$35,000 and \$200,000 and the study-region scenarios at \$75,000, we committed our predictions for the income and study-region tests to version control. Of 16,200 votes, two failed to parse.

\textbf{Measuring WTP.} For each model, scenario, and income, we score WTP as the first tax at which fewer than half the votes are yes (corresponding to a nonparametric estimate of the median WTP), a rule fixed before the second round. When the vote share stays at or above one-half through \$3,000, WTP is censored; comparisons in which both sides are censored are left unscored. As a second measure, we fit the standard logit for referendum data, $\Pr(\text{yes} \mid t) = 1 / [1 + \exp(-(\alpha - \beta t))]$, which gives a fitted median ($=$ mean) WTP $= \alpha/\beta$ \citep{hanemann1984}, estimated by penalized likelihood \citep{firth1993}. The logit estimates are in Appendix Table~\ref{tab:logit}.

\section{Results}

\subsection{Theoretical validity}

In this application, theoretical validity takes the form of three predictions from economic theory. Each stands for a general principle: answers should respond to cost, to how much of the good is on offer, and to the resources of the person advised.

\begin{figure}[ht]
\centering
\includegraphics[width=0.9\textwidth]{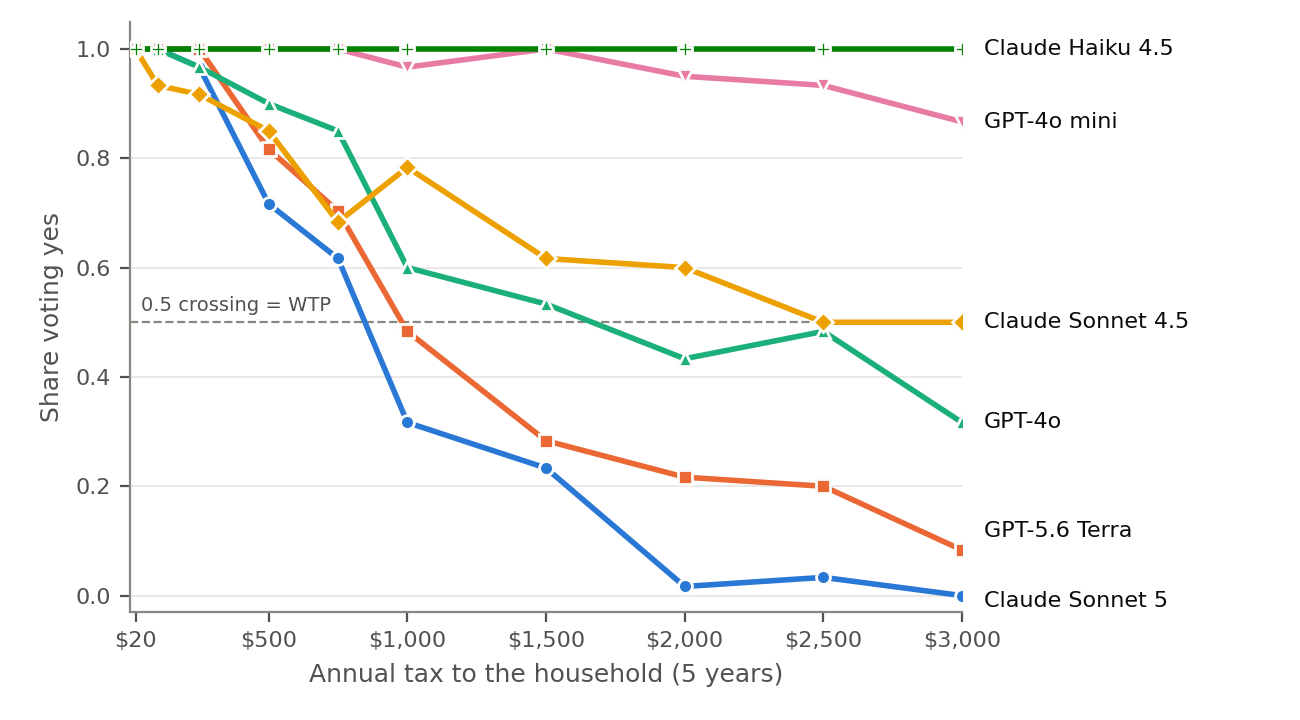}
\caption{Share of yes votes by annual tax at a \$75,000 household income, pooled over the six watershed scenarios (60 draws per point). A valid valuation requires a downward slope.}
\label{fig:yesshare}
\end{figure}

\textbf{Demand slopes down.} The most basic test is whether the share voting yes falls as the tax rises. Some models generate downward-sloping curves; others are flat (Figure~\ref{fig:yesshare}). Claude Haiku 4.5 votes yes in all 600 of its first-round draws, including at \$3,000 a year for five years on a \$75,000 income. GPT-4o mini still votes yes 87\% of the time at \$3,000. Neither yields a recoverable WTP in most scenarios. Claude Sonnet 5 and GPT-5.6 Terra trace clear demand curves, falling to 0.00 and 0.08 yes at \$3,000. Claude Sonnet 4.5 and GPT-4o respond to price for the non-local watershed but vote yes almost regardless of price for their own. The split is by generation and size, not developer.

\textbf{Scope.} WTP should respond to how much of the good is delivered. The sharpest test is nested dominance: minimum Level 2 contains minimum Level 3, so its WTP should be at least as high. There is one violation across all models, from Haiku at the lowest income, within noise. Scope also has a spatial dimension. WTP should fall for a watershed that does not contain the home (distance decay), and should not fall when the region grows to contain the home watershed and more (spatial extent). Distance decay holds in all 33 scored comparisons. The spatial-extent test separates the models: Sonnet 5 and Terra pass every scored comparison, while Sonnet 4.5 and GPT-4o price the full region below the local watershed it contains in five of eight.

\textbf{Income.} WTP should rise with income. It never falls in any of 69 scored comparisons, although much of the \$200,000 column is censored above \$3,000.

\subsection{Convergent validity}

Convergent validity asks whether different measures of the same thing agree. \textbf{Across estimators}, they do: re-scoring the tests with the logit estimates produces no significant violation of nested dominance, distance decay, or income. Since all estimators use the same raw data, this is a weak test of convergent validity. For spatial extent, Sonnet 4.5's failure on minimum Level 3 at \$35,000 is significant (region \$919 vs.\ local \$1,620), while GPT-4o's are not, nor is a reversal in Sonnet 5's point estimates at \$75,000 (Table~\ref{tab:logit}). \textbf{Across elicitation formats}, they do less well. In an earlier round we asked the same nine scenarios as an open-ended question about the maximum tax, with the four older models at a \$100,000 income. The formats disagree most for the models that the referendum could not score. Asked for a maximum, Haiku's median answers range from \$575 to \$1,650 a year and GPT-4o mini's from \$400 to \$1,200, every cell below \$3,000. Yet in the referendum, at a lower income, Haiku votes yes at \$3,000 in every draw and GPT-4o mini in 87\%. The open-ended answers were also sensitive to wording within that format: letting the model reason before giving its figure, instead of answering with a bare dollar amount, multiplied Haiku's and GPT-4o's answers by 2.8 and 2.3, a reliability problem as much as a convergent one. \textbf{Against the human estimates}, the newest models' levels are high. At \$75,000, the logit estimates for Sonnet 5 average about four times the human means and those for Terra about six times. They also scale with income faster than people do: between \$35,000 and \$75,000 their estimates imply income elasticities of WTP between 1.0 and 2.2, where SP studies usually find well below one \citep{kristrom1996}. They pass the income test in sign but not in magnitude.

\subsection{Reliability}

Because each vote is an independent draw, reliability can be read from sampling variation. With ten draws per tax the standard error of a vote share near one half is about 0.15, and most upticks in the vote curves are within that. Reliability across model versions is harder to establish: the two newest models are available only under undated names, so a later run under the same name may reach a different model.

\section{Discussion}

\textbf{The framework discriminates, and says where.} Applied to one instrument, the validity framework does more than rank models. It locates their failures. The oldest models fail the most basic theoretical test, and for them the other tests are empty. The newest pass every theoretical test we can score and diverge only under convergent validity, in the level and income sensitivity of their values. A single accuracy score could not make that distinction, and without ground truth there would be no accuracy score to compute.

\textbf{Consistency is not accuracy.} A model that passes these tests has shown that its answers are coherent, not that they are right. Theoretical validity is a necessary condition for trusting a model's judgement, not a sufficient one. A model could also pass by applying a well-learned rule of thumb, such as a fixed share of income discounted for distance, and these tests cannot tell such a rule from genuine weighing of the scenario. Models trained on text that includes the SP literature may know what the tests expect. Withholding the scale's name limits that risk; it does not remove it.

\textbf{What does not transfer is where the open questions are.} Criterion validity has no direct counterpart, because a model never pays; its analogue, whether what a model says matches what it does when acting, is a question for agentic evaluation. Incentive compatibility and consequentiality presuppose a respondent with something at stake. Whether an LLM's answers change when it is told its advice will be acted on, or when a question's format invites agreement, is testable, and it connects SP practice directly to current concerns about evaluation awareness and sycophancy.

\textbf{Limits.} This is one instrument and one good. The tax ladder tops out at \$3,000, so many comparisons pass by default. Ten draws per tax make individual calls noisy, and the first round was exploratory. The two newest models ran under undated names at provider-default settings. At \$75,000, the local- and non-local-watershed cells come from the first round and the study-region cells from the second, a week later. Because Sonnet 5 and Terra were called under undated names, the two sides of those spatial-extent comparisons, and of the \$35,000-to-\$75,000 income comparisons for the watershed cells, may not come from the same model. The same-day \$35,000 comparisons give the same spatial-extent verdicts: Sonnet 5 and Terra pass all three, and four of the five failures occur there. We did not swap the order of the Yes and No options (the prompt lists No first). A yes-bias tied to option order or wording would raise vote shares at every tax. For models whose curves fall within the tax range this would mostly shift the level, but for the models that vote yes at nearly every price it could be what makes the curve flat. We plan to run the swap.

\textbf{Conclusion.} The SP literature spent decades learning how to judge answers that cannot be checked against the truth. Those lessons carry over to language models, whose answers to questions of value, judgement, and advice cannot be checked against the truth either. The next steps are to apply the framework outside economics and to test the concepts that do not yet transfer.

\section*{Appendix}
\addcontentsline{toc}{section}{Appendix}

\renewcommand{\thetable}{A\arabic{table}}
\setcounter{table}{0}

\begin{table}[H]
\centering
\footnotesize
\setlength{\tabcolsep}{3.2pt}
\renewcommand{\arraystretch}{1.1}
\begin{tabular}{@{}l *{9}{r}@{}}
\toprule
 & \multicolumn{3}{c}{\textbf{Local watershed}} & \multicolumn{3}{c}{\textbf{Non-local watershed}} & \multicolumn{3}{c}{\textbf{Full study region}} \\
\cmidrule(lr){2-4} \cmidrule(lr){5-7} \cmidrule(l){8-10}
\textbf{Model} & \textbf{One-level} & \textbf{Min L2} & \textbf{Min L3} & \textbf{One-level} & \textbf{Min L2} & \textbf{Min L3} & \textbf{One-level} & \textbf{Min L2} & \textbf{Min L3} \\
\midrule
Claude Sonnet 5 & 1,155 & 1,808 & 1,220 & 462 & 651 & 485 & 1,463 & 1,753 & 1,058 \\
 & (90) & (114) & (100) & (53) & (51) & (54) & (119) & (105) & (68) \\[2pt]
GPT-5.6 Terra & 1,623 & 3,156$^\dagger$ & 1,184 & 925 & 801 & 541 & 2,066 & 3,156$^\dagger$ & 1,747 \\
 & (135) & (301) & (100) & (95) & (74) & (60) & (158) & (301) & (134) \\[2pt]
GPT-4o & 3,156$^\dagger$ & -- & 2,945 & 936 & 957 & 617 & 2,524 & -- & 2,657 \\
 & (301) &  & (221) & (88) & (73) & (67) & (139) &  & (162) \\[2pt]
Claude Sonnet 4.5 & -- & -- & -- & 677 & 1,280 & 949 & -- & -- & 3,143$^\dagger$ \\
 &  &  &  & (116) & (147) & (117) &  &  & (240) \\[2pt]
GPT-4o mini & -- & -- & -- & -- & 3,439$^\dagger$ & 3,329$^\dagger$ & -- & -- & -- \\
 &  &  &  &  & (479) & (453) &  &  &  \\[2pt]
Claude Haiku 4.5 & -- & -- & -- & -- & -- & -- & -- & -- & -- \\
\addlinespace[2pt]
\midrule
\emph{Human mean} & \emph{316} & \emph{492} & \emph{217} & \emph{165} & \emph{225} & \emph{95} & \emph{300} & \emph{463} & \emph{207} \\
 & \emph{(13)} & \emph{(21)} & \emph{(10)} & \emph{(11)} & \emph{(15)} & \emph{(9)} & \emph{(12)} & \emph{(18)} & \emph{(9)} \\
\bottomrule
\end{tabular}
\caption{Estimated WTP at a \$75,000 household income: $\alpha/\beta$ from a Firth-penalized logit of the vote on the tax, fitted to 100 votes per cell, dollars per year for five years; delta-method standard errors in parentheses. ``--'' means the price coefficient is not significantly positive (a flat curve). $^\dagger$ marks an estimate above the \$3,000 top bid. Human means are from \citet{vossler2023}, Model 1 (mixed logit), in 2021 dollars, and average over respondents' own watersheds, which vary in size; the comparison is approximate.}
\label{tab:logit}
\end{table}

\bibliographystyle{apalike}
\bibliography{refs}

\end{document}